\documentclass[letterpaper, 10 pt, conference]{ieeeconf}  

\IEEEoverridecommandlockouts                              

\usepackage{amsmath}
\usepackage{amsfonts}
\usepackage{cases}
\usepackage{amssymb}
\usepackage{booktabs}
\usepackage{bigstrut}
\usepackage{graphicx}
\usepackage{subcaption}
\usepackage{caption}
\usepackage{cite}
\usepackage{bigstrut}
\usepackage{url}

\title{\LARGE \bf
Whole-Body Control Framework for Humanoid Robots with \\ Heavy Limbs: A Model-Based Approach
}

\author{Tianlin Zhang, Linzhu Yue, Hongbo Zhang, Lingwei Zhang, \\ Xuanqi Zeng, Zhitao Song and Yun-Hui Liu
\thanks{
T. Zhang, L. Yue, H. Zhang, L. Zhang, X. Zeng, Z. Song  and Y. Liu with the Department of Mechanical and Automation Engineering at the Chinese University of Hong Kong, Hong Kong SAR, China. {\tt\small skywoodszcn@gmail.com}}%
\thanks{* Corresponding author: Y. Liu {\tt\small yhliu@cuhk.edu.hk}}
\thanks{This work is supported by the InnoHK Clusters of the Hong Kong SAR Government via the Hong Kong Centre for Logistics Robotics, and the CUHK T Stone Robotics Institute.
}
}

\begin{document}

\maketitle
\thispagestyle{empty}
\pagestyle{empty}

\begin{abstract}
Humanoid robots often face significant balance issues due to the motion of their heavy limbs. These challenges are particularly pronounced when attempting dynamic motion or operating in environments with irregular terrain.
To address this challenge, this manuscript proposes a whole-body control framework for humanoid robots with heavy limbs, using a model-based approach that combines a kino-dynamics planner and a hierarchical optimization problem.
The kino-dynamics planner is designed as a model predictive control (MPC) scheme to account for the impact of heavy limbs on mass and inertia distribution. By simplifying the robot's system dynamics and constraints, the planner enables real-time planning of motion and contact forces.
The hierarchical optimization problem is formulated using Hierarchical Quadratic Programming (HQP) to minimize limb control errors and ensure compliance with the policy generated by the kino-dynamics planner.
Experimental validation of the proposed framework demonstrates its effectiveness. The humanoid robot with heavy limbs controlled by the proposed framework can achieve dynamic walking speeds of up to 1.2~m/s, respond to external disturbances of up to 60~N, and maintain balance on challenging terrains such as uneven surfaces, and outdoor environments.
\end{abstract}

\section{INTRODUCTION}
\begin{figure}[tp]
  \centering
  \includegraphics[width=0.4\textwidth]{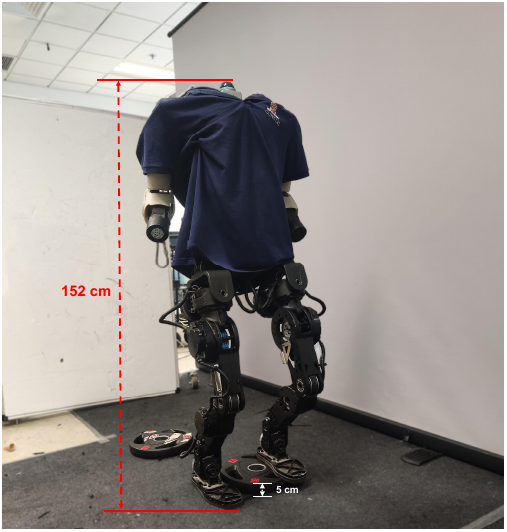}
  \caption{Illustration of a humanoid robot with two 6-DoF manipulators and two 6-DoF legs, walking over uneven terrain.}
  \label{fig:cover}
\end{figure}

Humanoid robots, shown in Fig.~\ref{fig:cover}, have vast potential applications due to their ability to adapt to and interact with human-centric environments, such as homes, workplaces, and healthcare settings\cite{gu2025humanoid}. However, to achieve sufficient actuation and accuracy, humanoid robots are typically equipped with servomotors and gearboxes, which result in a high mass ratio in their limbs. Specifically, the limbs often account for 40\% to 60\% of the total mass in humanoid robots, such as Digit\cite{castillo2021robust} and Unitree G1\cite{wang2020unitree}. This mass distribution makes the coupling between the base and limbs a critical factor for the robot's balance, presenting significant challenges in controller design.

One issue arising from the base-limbs coupling is that the robot's mass and inertia distribution are not solely determined by the base but also change with the motion of the limbs. As a result, the center of mass (CoM) of the robot experiences significant fluctuations due to the rapid or large-range motion of the limbs, which can induce instability if not properly accounted for. Reducing the speed and range of limb motion is one way to mitigate this risk. However, this approach leads to poor responsiveness to external disturbances and decreased operational efficiency, making it unsuitable for many applications.
Another challenge is that the robot's balance is affected by control errors in the limbs. This becomes particularly problematic during dynamic tasks (e.g., rapid motion and responding to external disturbances), where such errors can cause the base to deviate from its desired trajectory, leading to overshooting, oscillations, and other unstable behaviors. This is because control errors in the limbs are magnified by the inertia effects and transferred to the base, making it difficult to maintain balance.

A broad range of work aimed at addressing this problem can be categorized into learning-based and model-based approaches, both of which decompose the full control problem into two main components: a planning module and a tracking module.
In learning-based methods, the planning module typically generates the robot's reference motion using reinforcement learning\cite{gu2024humanoid} or imitation learning\cite{ji2024exbody2}. This approach implicitly captures the base-limbs coupling through data-driven techniques, eliminating the need for explicit physical modeling. However, a key drawback of this method is its lack of interpretability, as the learned policies are often treated as black-box models, making it difficult to understand or predict the robot's behavior in real-world scenarios\cite{li2019formal}. The tracking module in learning-based methods typically employs high-gain Proportional-Integral-Derivative (PID) controllers to generate torque commands for tracking the reference motion. However, due to the nonlinear dynamics of humanoid robots, the performance of the PID controller degrades, making it challenging to effectively reduce limb control errors.
In model-based methods, the planning module generates the robot's reference motion and contact forces by explicitly modeling the base-limbs coupling. It typically formulates a model predictive control (MPC) scheme based on the full kinematics and dynamics with high fidelity (e.g., centroidal dynamics\cite{orin2013centroidal} and full dynamics\cite{mason2014full}), incorporating multiple constraints to ensure physically feasible and safe motion in real-world scenarios\cite{scianca2020mpc}. However, the high dimensionality of the optimization variables and the large number of constraints in the MPC make real-time computation challenging\cite{dai2014whole,dadiotis2023whole}. The tracking module in model-based methods consists of an instantaneous feedback controller that accounts for the full dynamics of the robot. By considering the robot's nonlinear dynamics, this controller achieves better tracking performance compared to the PID controller\cite{kim2019highly,wensing2023optimization,zhang2024whole}. However, the instantaneous feedback controller operates independently with different control objectives and constraints and, therefore, does not explicitly ensure compliance with the policy generated by the planner, as argued in \cite{mastalli2022agile}. Consequently, this discrepancy can make it difficult to eliminate limb control errors, causing the robot to lose stability.

To address these challenges, this manuscript proposes a whole-body control framework for humanoid robots with heavy limbs using a model-based approach. The framework consists of a kino-dynamics planner and a hierarchical optimization problem. The kino-dynamics planner reduces the computational complexity of the MPC by simplifying the system dynamics and constraints. This simplification enables the planner to generate reference motion and contact forces in real-time while accounting for the impact of heavy limbs on the robot's mass and inertia distribution. The hierarchical optimization problem is formulated using hierarchical quadratic programming (HQP) to prioritize reference motion and contact forces tracking, minimize limb control errors, and ensure compliance with the policy generated by the planner. 

This work offers the following contributions:
\begin{itemize}
  \item This paper proposes a novel whole-body control framework for humanoid robots with heavy limbs through a model-based approach. The proposed framework effectively eliminates the negative effects caused by base-limbs coupling, achieves speeds of up to 1.2~m/s, resists external disturbances of up to 60~N, and maintains stability while walking on challenging terrains.
  \item The proposed framework not only operates in real-time but also minimizes limb tracking errors, ensuring compliance with the policy generated by the planner.
  \item The proposed method has been validated through a series of experiments and compared with previous approaches.
\end{itemize}

\begin{figure}[tp]
  \centering
  \includegraphics[width=0.35\textwidth]{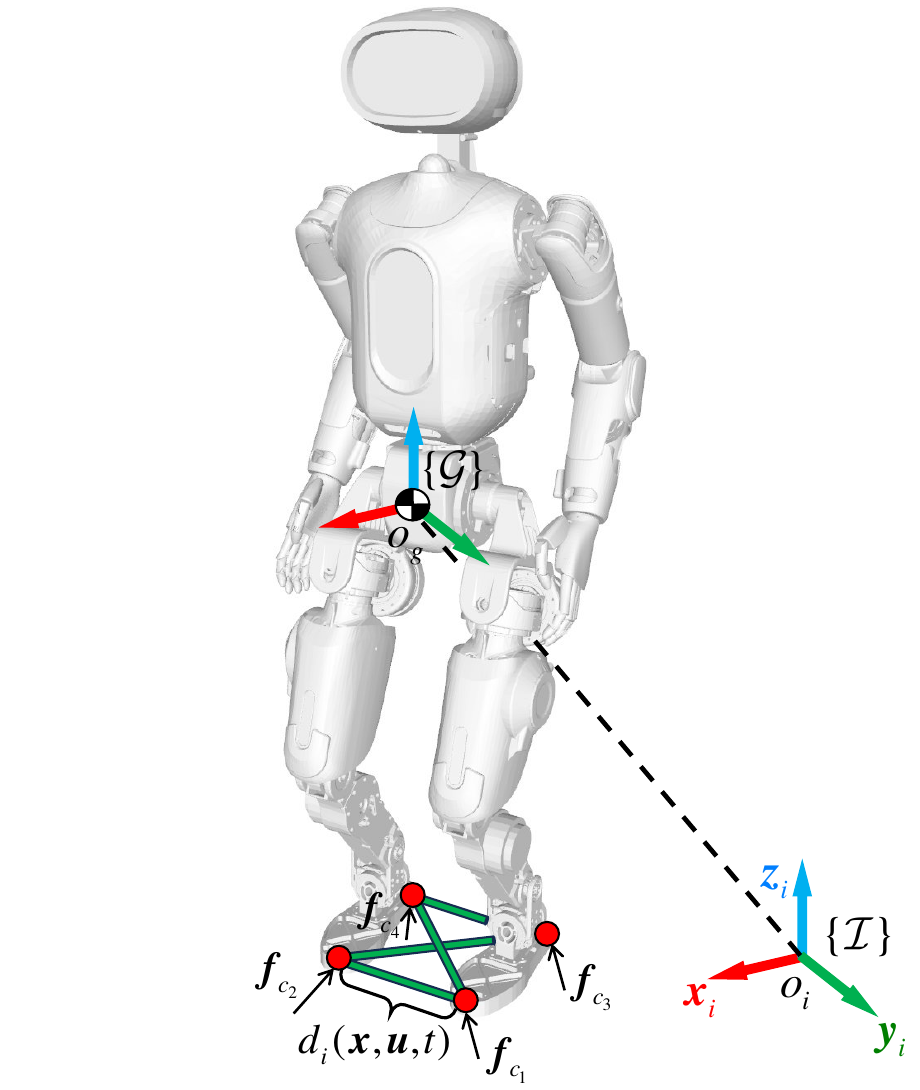}
  \caption{Illustration of the humanoid robots frames (inertial $\mathcal{I}$ and centroidal $\mathcal{G}$), contact forces $\boldsymbol{f}_{c_i}$, and the collision constraint ${d_i}({\boldsymbol{x}},{\boldsymbol{u}},t)$.}
  \label{fig:system}
\end{figure}

\section{WHOLE-BODY CONTROL FRAMEWORK}
\subsection{Kino-Dynamics Planner}\label{Sec:MPC}
Inspired by the work of \cite{sleiman2021unified}, the kino-dynamics planner is built using the MPC scheme, which includes system dynamics ${\boldsymbol{\dot x}}(t) = {\boldsymbol{f}}({\boldsymbol{x}}(t),{\boldsymbol{u}}(t),t)$, equality constraints ${\boldsymbol{g}_i({\boldsymbol{x}(t)},{\boldsymbol{u}(t)},t) = {\boldsymbol{0}}}$, and inequality constraints ${\boldsymbol{h}_i({\boldsymbol{x}(t)},{\boldsymbol{u}(t)},t) \geqslant \boldsymbol{0}}$.

\subsubsection{System dynamics}
The system dynamics of the humanoid robot is described using the two frames: the right-hand inertial coordinate frame $\left\{ {\mathcal{I}:{o_i} - {{\boldsymbol{x}}_i}{{\boldsymbol{y}}_i}{{\boldsymbol{z}}_i}} \right\}$ and the centroidal frame $\left\{ {\mathcal{G}:{o_g} - {{\boldsymbol{x}}_g}{{\boldsymbol{y}}_g}{{\boldsymbol{z}}_g}} \right\}$, as shown in Fig.~\ref{fig:system}. 
The centroidal frame $\mathcal{G}$ is a reference frame attached to the robot's CoM and aligned with the inertial frame $\mathcal{I}$.
To account for the impact of limb motion on the robot's mass and inertia distribution, a full kinematics with centroidal dynamics is used to describe the humanoid robot, as follows:
\begin{subequations}
\label{eq:cd}
    \begin{align}
        &\dot{\boldsymbol{h}}_{com} = \left[ {\begin{array}{*{20}{c}}
          {\sum\limits_{i = 1}^{{n_c}} {{{\boldsymbol{f}}_{{c_i}}}}  + m{\boldsymbol{g}}} \\ 
          {\sum\limits_{i = 1}^{{n_c}} {{{\boldsymbol{r}}_{com,{c_i}}}}  \times {{\boldsymbol{f}}_{{c_i}}}} 
        \end{array}} \right], \label{rq:cd_1}\\
        &\dot{\boldsymbol{q}}_b = {\boldsymbol{A}}_b^{- 1}\left( {{{\boldsymbol{h}}_{com}} - {{\boldsymbol{A}}_j}{{\boldsymbol{v}}_j}} \right), \label{rq:cd_2}\\
        &\dot{\boldsymbol{q}}_j = {{\boldsymbol{v}}_j} \label{rq:cd_3},
    \end{align}
\end{subequations}
where $\boldsymbol{h}_{com} \in \mathbb{R}^6$ represents the centroidal momentum about the centroidal frame $\mathcal{G}$, including both the linear and angular momentum, $m\boldsymbol{g}$ represents the gravity of the robot, $\boldsymbol{r}_{com,{c_i}} \in \mathbb{R}^3$ represents the position of the contact points with respect to CoM, $\boldsymbol{f}_{c_i} \in \mathbb{R}^3$ is the contact forces at the $i$th contact point, respectively, ${\boldsymbol{q}},{\boldsymbol{v}} \in {\mathbb{R}^{6 + {n_j}}}$ are the stacked vector of generalized coordinates and generalized velocities, respectively, a ZYX-Euler angle parameterization is assumed to represent the base’s orientation, the matrix $\boldsymbol{A}$ represents the centroidal momentum matrix, the subscripts $b$ and $j$ stand for the base and actuated limbs, respectively, $n_c$ and $n_j$ represents the number of contact points and degrees of freedom (DoF) of actuated limbs, respectively. 
From \eqref{eq:cd}, the relationship between the limb motion and the robot’s balance is established through the centroidal momentum, as detailed in \cite{orin2013centroidal}.

Compared to previous studies \cite{li2021force,garcia2021mpc,li2023dynamic}, \eqref{eq:cd} assumes that the robot has only four possible contact points and ignores the contact momentums acting on the feet, as shown in Fig.~\ref{fig:system}. Although this simplification results in underactuation of the humanoid robot's base during single-leg support, the prediction provided by the MPC helps mitigate the negative effects of underactuation, as discussed in \cite{di2018dynamic}. More importantly, this simplification reduces the dimensionality of the optimization variables and eliminates the need for additional contact wrench consistency (CWC) constraints \cite{caron2015stability}, making real-time computation feasible.

With the simplified system dynamics \eqref{eq:cd}, the state vector of MPC is chosen as  ${\boldsymbol{x}} = \left[ {{\boldsymbol{h}}_{com}^T,{\boldsymbol{q}_b^T},{\boldsymbol{q}_j^T}} \right]^T \in {\mathbb{R}^{12 + {n_j}}}$, and the input vector is chosen as ${\boldsymbol{u}} = \left[ {{\boldsymbol{f}}_{{c_1}}^T,...,{\boldsymbol{f}}_{{c_4}}^T,{\boldsymbol{v }}_j^T} \right]^T    \in {\mathbb{R}^{12 + {n_j}}}$.

\subsubsection{Equality constraints}
Equality constraints are defined based on the contact states, which can be either open or closed. Let $\mathcal{C}$ denote the set of closed contacts.  When the foot makes contact with the ground (i.e., ${{c_i} \in \mathcal{C}}$), the planner ensures that the foot remains stationary relative to the ground. Conversely, when the foot loses contact with the ground (i.e., ${{c_i} \in \overline{\mathcal{C}}}$), the planner generates the reference swing trajectory of the foot in operational space based on the upcoming footstep location. Therefore, the equality constraints can be established as follows:
\begin{subequations}
\begin{numcases}{}
    \boldsymbol{v}_{c_i} = \boldsymbol{0}, \ \ \ \ \ \ \ \ if \ c_i \in \mathcal{C}, \\
    \boldsymbol{f}_{c_i} = \boldsymbol{0}, \ \ \ \ \ \ \ \ if \ c_i \in \overline{\mathcal{C}},\\ 
    \boldsymbol{p}_{c_i}(t) = \boldsymbol{p}_{c_i}^*(t), \ \ \ if \ c_i \in \overline{\mathcal{C}},\\
    \dot{\boldsymbol{p}}_{c_i}(t) = \dot{\boldsymbol{p}}_{c_i}^*(t), \ \ \ if \ c_i \in \overline{\mathcal{C}},
\end{numcases}
\end{subequations}
where ${{\boldsymbol{p}}_{{c_i}}(t)},{\dot{\boldsymbol{p}}_{{c_i}}(t)} \in {\mathbb{R}^3}$ are the position and linear velocity of the foot, expressed in the inertial frame $\mathcal{I}$, respectively, ${{\boldsymbol{p}}_{c_i}^*}(t),{\dot{\boldsymbol{p}}_{c_i}^*}(t) \in {\mathbb{R}^3}$ represent the desired position and velocity of the foot described by the cubic trajectory generated from the upcoming footstep location, respectively. The upcoming footstep location ${\boldsymbol{p}}_{{c_i}}^d \in {\mathbb{R}^3}$ is calculated from the corresponding hip location using a combination of the Raibert heuristic, a velocity-based feedback term from the capture point and centrifugal formulation\cite{raibert1986legged} as follows: 
\begin{equation}
\label{eq:desired_swing_traj}
{\boldsymbol{p}}_{{c_i}}^d = {{\boldsymbol{p}}_{hi{p_{{c_i}}}}} + {{\boldsymbol{p}}_{capture}} + {{\boldsymbol{p}}_{centrifugal}},
\end{equation}
where ${{\boldsymbol{p}}_{hi{p_{{c_i}}}}} \in {\mathbb{R}^3}$ is the position of the corresponding hip, ${{\boldsymbol{p}}_{capture}}$ and  ${{\boldsymbol{p}}_{centrifugal}} \in {\mathbb{R}^3}$ are calculated as follows:
\begin{subequations}
\begin{align}
&{{\boldsymbol{p}}_{capture}} = \frac{{{\Delta t}}}{2}{\dot{\boldsymbol{p}}}_b^d + k({\dot{\boldsymbol{p}}_b} - {\dot{\boldsymbol{p}}}_b^d), \\
&{{\boldsymbol{p}}_{centrifugal}} = \sqrt {\frac{h}{\left|\boldsymbol{g}\right|}} {\dot{{\boldsymbol{p}}}_b} \times {\boldsymbol{\omega }}_b^d,
\end{align}
\end{subequations}
where $\Delta t$ represents the time duration of the contact phase (stance time), ${\dot{\boldsymbol{p}}}_b^d$ and ${\dot{\boldsymbol{p}}_b} \in {\mathbb{R}^3}$ represent the desired and actual linear velocity of the base, respectively, $h$ denotes the nominal height of the base, and ${\boldsymbol{\omega}}_b^d$ is the desired angular velocity of the base, $k \in \mathbb{R}$ is the velocity feedback gain, which is set to 0.15 in this manuscript.

\subsubsection{Inequality constraints}
To ensure that the planner generates dynamically feasible motions and forces, inequality constraints are established to respect the system’s intrinsic operational limits and prevent self-collision, as follows:
\begin{subequations}
\label{eq:ic}
\begin{numcases}{}
    {{\mu}_s}f_{c_i}^z - \sqrt {f_{{c_i}}^{{x^2}} + f_{{c_i}}^{{y^2}}} \geqslant 0, \ \ \ \ \ if \ {c_i} \in \mathcal{C}, \label{eq:ic_friction_cone} \\
    {0 < f_{c_i}^z \leqslant f_{max}^z}, \ \ \ \ \ \ \ \ \ \ \ \ \ \ \ \ if \ {c_i} \in \mathcal{C}, \label{eq:ic_fz_max} \\\
    {{\boldsymbol{q}}_{{j_{min}}}} \leqslant {{\boldsymbol{q}}_j} \leqslant {{\boldsymbol{q}}_{{j_{max}}}}, \label{eq:ic_joint_pos_max}\\
    -{{\boldsymbol{v}}_{{j_{max }}}} \leqslant {{\boldsymbol{v}}_j} \leqslant {{\boldsymbol{v}}_{{j_{max }}}}, \label{eq:ic_joint_vel_max}\\
    {d_i}({\boldsymbol{x}},\boldsymbol{u},t) \geqslant {\varepsilon _i}, \label{eq:ic_collision}
\end{numcases}
\end{subequations}
where \eqref{eq:ic_friction_cone} ensures that the contact forces in the $x$ and $y$ directions remain within the friction pyramid, with $\mu$ being the friction coefficient, 
\eqref{eq:ic_fz_max} enforces that the contact forces in the $z$-direction stay within the defined upper and lower force bounds, \eqref{eq:ic_joint_pos_max} and \eqref{eq:ic_joint_vel_max} impose constraints on joint position and velocity limits, ensuring they are not violated, \eqref{eq:ic_collision} describes collision avoidance, where ${d_i}({\boldsymbol{x}},\boldsymbol{u},t)$ represents the distance between a collision pair and ${\varepsilon _i}$ denotes the minimum allowed distance threshold for each collision pair.

Since humanoid robots are typically designed with a slender body structure, the small distances between links make self-collisions more likely. Such collisions can damage the robot's hardware and make the robot unstable. To address this issue, previous studies have commonly enclosed each link within a spherical approximation and prevented self-collisions by constraining the distances between these spheres \cite{chiu2022collision,khazoom2024tailoring,schwienbacher2011self}. However, the computational complexity of this method increases quadratically with the number of links, making real-time computation infeasible for multi-link humanoid robots.
To address this challenge, the collision model is simplified, as shown in Fig.~\ref{fig:system}. 
The collision model for the robot's lower limbs only considers the distance between the right and left feet, as follows:
\begin{equation}
\label{eq:collsion}
  {d_i}({\boldsymbol{x}},\boldsymbol{u},t) = {\left\| {{{\boldsymbol{p}}_{{c_i}}} - {{\boldsymbol{p}}_{{c_j}}}} \right\|^2},
\end{equation}
where $i,j \in \left\{ {1,...,{n_c}} \right\},i \ne j$. \eqref{eq:collsion} prevents foot from crossing the virtual plane connected to the other foot to avoid the collision. 
The collision between the upper limbs is constrained by joint position limits \eqref{eq:ic_joint_pos_max}.
For most applications, joint limits effectively prevent self-collisions in links other than the feet (e.g., the upper limbs or thighs). By combining the joint limits in \eqref{eq:ic_joint_pos_max} with the feet collision constraints in \eqref{eq:collsion}, the planner can prevent self-collisions while reducing unnecessary computational load, thus improving real-time performance.

\begin{figure*}[tp]
  \centering
  \includegraphics[width=1.0\textwidth]{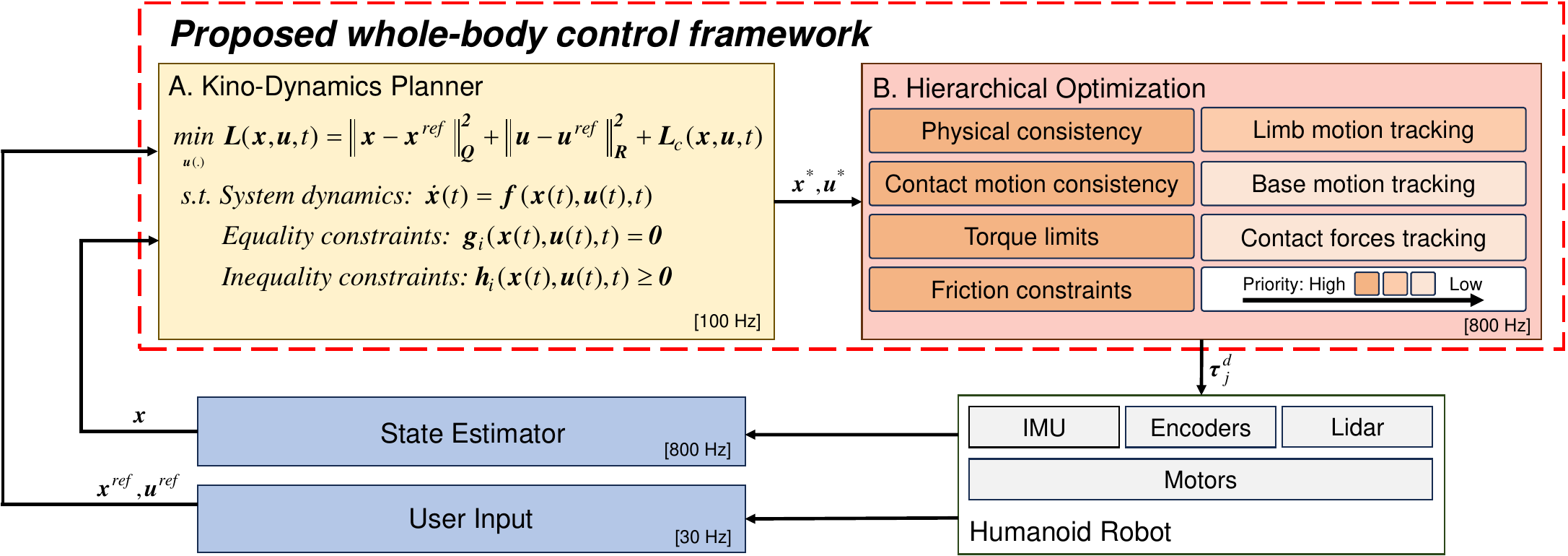}
  \caption{The block diagram illustrates the proposed framework for a humanoid robot with heavy limbs. First, the kino-dynamics planner generates reference motion and forces ($\boldsymbol{x}^*$ and $\boldsymbol{u}^*$) based on the user input ($\boldsymbol{x}^{ref}$ and $\boldsymbol{u}^{ref}$) and state estimator.
  Then, a hierarchical optimization problem is formulated to track $\boldsymbol{x}^*$ and $\boldsymbol{u}^*$, generating torque commands ${\boldsymbol{\tau }}_j^d$ for the humanoid robot's actuators.
  }
  \label{fig:framework}
\end{figure*}

\subsubsection{Cost}
The planner's cost function is defined as follows:
\begin{equation}
    \boldsymbol{L}({\boldsymbol{x}},{\boldsymbol{u}},t) = \left\| {{\boldsymbol{x}} - {{\boldsymbol{x}}^{ref}}} \right\|_{\boldsymbol{Q}}^2 + \left\| {{\boldsymbol{u}} - {{\boldsymbol{u}}^{ref}}} \right\|_{\boldsymbol{R}}^2 + \boldsymbol{L}_c({\boldsymbol{x}},{\boldsymbol{u}},t),
\end{equation}
which includes a state tracking cost term $\left\| {{\boldsymbol{x}} - {{\boldsymbol{x}}^{ref}}} \right\|_{\boldsymbol{Q}}^2$, an input regularization term $\left\| {{\boldsymbol{u}} - {{\boldsymbol{u}}^{ref}}} \right\|_{\boldsymbol{R}}^2$ and an inequality constraint penalty term $\boldsymbol{L}_c({\boldsymbol{x}},{\boldsymbol{u}},t)$. 

To enforce the inequality constraints \eqref{eq:ic} in real-time, all inequality constraints are converted into penalty costs through Radial Basis Functions (RBF)\cite{grandia2019feedback}, as follows:
\begin{equation}
\label{eq:rbf}
  \boldsymbol{B}(\boldsymbol{h}) = \left\{ {\begin{array}{*{20}{r}}
  { - \xi \ln (\boldsymbol{h}),}&{\boldsymbol{h} \geqslant \delta }, \\ 
  {\xi \boldsymbol{\beta} (\boldsymbol{h};\delta ),}&{\boldsymbol{h} < \delta }, 
\end{array}} \right. 
\end{equation}
which is defined as a log-barrier on the interior of the feasible space and switches to a quadratic function at a distance $\delta \in \mathbb{R}$ from the constraint boundary. 
Using the RBF \eqref{eq:rbf},  the inequality constraint penalty term is:
\begin{equation}
\boldsymbol{L}_c\left( {{\boldsymbol{x}},{\boldsymbol{u}},t} \right) = \sum\limits_{i = 1}^{{n_p}} \boldsymbol{B} \left( {\boldsymbol{h}_i\left( {{\boldsymbol{x}},{\boldsymbol{u}},t} \right)} \right), 
\end{equation}
where $n_p$ is the total number of the inequality constraints. In this way, the problem is transformed into an optimization problem with only equality constraints, simplifying the calculation \cite{khazoom2024tailoring}.

\begin{table}[tp]
  \centering
  \caption{THE TASKS USED IN THE HIERARCHICAL OPTIMIZATION. EACH TASK IS ASSOCIATED WITH A PRIORITY (1 IS THE HIGHEST).}
    \begin{tabular}{cl}
    \hline
    Priority & Task \bigstrut \\
    \hline
    1     & Physical consistency \bigstrut[t]\\
          & Contact motion consistency \\
          & Torque limits \\
          & Friction constraints \\
    2     & Limb motion tracking \\
    3     & Base motion tracking \\
    4     & Contact forces tracking \bigstrut[b]\\
    \hline
    \end{tabular}%
  \label{tab:wbc}%
\end{table}%

\subsection{Hierarchical Optimization}\label{Sec:WBC}
The hierarchical optimization problem based on HQP \cite{bellicoso2016perception} is formulated to track the reference motion and forces generated by the kino-dynamics planner (Sec.~\ref{Sec:MPC}). It prioritizes tracking limb motion over the base and contact forces reference, as shown in Tab.~\ref{tab:wbc}. This approach not only minimizes limb motion tracking errors but also prevents the robot from exploiting limb inertia to track the base references in underactuated directions. As a result, it ensures compliance with the policy generated by the planner, maintaining the robot’s stability and improving the overall performance of the humanoid robot.

Let ${\boldsymbol{\xi}} = {\left[ {{{\dot{\boldsymbol{v}}}^T},{\boldsymbol{F}}_c^T} \right]^T} \in {\mathbb{R}^{6 + {n_j} + 3{n_c}}}$ be the optimization variable, where $\dot{\boldsymbol{v}} \in \mathbb{R}^{6 + {n_j}}$ represents the generalized accelerations, and $\boldsymbol{F}_c = {\left[ {{\boldsymbol{f}}_{{c_1}}^T,...,{\boldsymbol{f}}_{{n_c}}^T} \right]^T} \in {\mathbb{R}^{3n_c}}$ is a stacked vector of contact forces. The solver of the hierarchical optimization problem searches for the optimal $\boldsymbol{\xi}_d$ in tasks and then generates the actuators’ torque commands.

\subsubsection{Physical consistency}
Since the simplified system dynamics \eqref{eq:cd} is used to build the real-time kino-dynamics planner (Sec.~\ref{Sec:MPC}), the reference motion and forces may be physically infeasible. Therefore, the motion and forces need to be adjusted using the full dynamics, as follows:
\begin{subequations}
    \label{eq:dynamics_full_body}
    \begin{align}
        &{{\boldsymbol{M}}_b}{\boldsymbol{\dot v}} + {{\boldsymbol{n}}_b} = {\boldsymbol{J}}_b^T{{\boldsymbol{F}}_c}, \label{eq:dynamics_full_body_base} \\
        &{{\boldsymbol {M}}_j}{\boldsymbol{\dot v}} + {{\boldsymbol{n}}_j} = {{\boldsymbol{\tau }}_j} + {\boldsymbol{J}}_j^T{{\boldsymbol{F}}_c},
        \label{eq:dynamics_joint}
    \end{align}
\end{subequations}
where $\boldsymbol{M}$ represents the inertia matrix, $\boldsymbol{n}$ denotes the nonlinear
effects (i.e., Coriolis, centrifugal, and gravitational terms), ${{\boldsymbol{\tau }}_j}$ is s the vector of actuators’ torques, ${\boldsymbol{J}}_b$ and ${\boldsymbol{J}}_j$ are matrices of the stacked contact jacobians, defined by the base and the actuated limbs, respectively.

To enhance real-time performance, the motion and contact forces are constrained to the manifold defined by the floating base part of dynamics \eqref{eq:dynamics_full_body_base}, as follows:
\begin{equation}
    \left[ {{{\boldsymbol{M}}_b}\ -{\boldsymbol{J}}_b^T} \right]{\boldsymbol{\xi }} =  - {{\boldsymbol{n}}_b}.
\end{equation}

\subsubsection{Limb motion tracking}
The limb motion tracking task is divided into two parts: swing leg motion and upper limb motion. For the swing legs, the motion is constrained in the operational space to track the reference swing trajectory as follows:
\begin{equation}
    \left[ {{{\boldsymbol{J}}_{{c}}}\ \boldsymbol{0}} \right]{\boldsymbol{\xi }} = k_p^c({{\boldsymbol{p}}_{{c}}} - {{\boldsymbol{p}}^*}(t)) + k_d^c({\dot{\boldsymbol{p}}_{{c}}} - {\dot{\boldsymbol{p}}^*}(t)) 
    -\dot{{\boldsymbol{J}}_{c}}{\boldsymbol{v}},
\end{equation}
where $k_p^c$ and $k_d^c \in \mathbb{R}$ are control gains for the position and linear velocity, respectively, ${{\boldsymbol{J}}_{{c}}}$ is the contact jacobian matrix.

For the upper limb, the motion is constrained in joint space to track the reference joint trajectory as follows: 
\begin{equation}
    \left[ {{{\boldsymbol{J}}_{{j}}}\ {\boldsymbol{0}}} \right]{\boldsymbol{\xi }} = k_p^j({{\boldsymbol{q}}_j} - {\boldsymbol{q}}_j^*) + k_d^j({{\dot{\boldsymbol{q}}}_j} - \dot{\boldsymbol{q}}_j^*) - \dot{{{\boldsymbol{J}}}_{{j}}}{\boldsymbol{v}},
\end{equation}
where $k_p^j$ and $k_d^j \in \mathbb{R}$ are control gains for the joints position and angular velocity, respectively.

\subsubsection{Base motion tracking}
The base reference motion is tracked in the operational space as follows:
\begin{equation}
    \left[ {{{\boldsymbol{J}}_{{b}}}\ {\boldsymbol{0}}} \right]{\boldsymbol{\xi}} = {\boldsymbol{K}}_p^b({{\boldsymbol{q}}_b} \boxminus {\boldsymbol{q}}_b^*) + {\boldsymbol{K}}_d^b({{\boldsymbol{v}}_b} - {\boldsymbol{v}}_b^*) - \dot{{{\boldsymbol{ J}}}_b}{\boldsymbol{v}},
\end{equation}
where ${\boldsymbol{q}}_b^*$ and ${\boldsymbol{v}}_b^* \in \mathbb{R}^6$ represent the reference base pose and velocity, respectively, the control gains ${\boldsymbol{K}}_p^b$ and ${\boldsymbol{K}}_d^b \in \mathbb{R}^{6 \times 6}$ are diagonal positive definite matrices, the box-minus operator $\boxminus$ \cite{bloesch2016primer} is used to compute the pose error.

\subsubsection{Contact forces tracking}
The priority of the contact forces tracking task is set to the lowest to regulate the force distribution in case of a contradiction between the reference motion and the forces. The contact forces tracking task is defined as follows:
\begin{equation}
    \left[ {{\boldsymbol{0}}\ {\boldsymbol{I}}} \right]{\boldsymbol{\xi}} = {\boldsymbol{F}}_c^*,
\end{equation}
where ${\boldsymbol{F}}_c^* \in {\mathbb{R}^{3n_c}}$ is the stacked vector of the reference contact forces.

\subsubsection{Contact motion consistency}
The solution found by the solver must ensure zero accelerations at the contact points of the stance legs. Therefore, the constraints can be calculated as follows:
\begin{equation}
  \left[ {{{\boldsymbol{J}}_{{c}}}\ {\boldsymbol{0}}} \right]{\boldsymbol{\xi }} =  - \dot{{{\boldsymbol{J}}}_{{c}}}{\boldsymbol{v}}.
\end{equation}

\subsubsection{Torque limits}
The actuation limits are enforced by utilizing the actuated part of the dynamics \eqref{eq:dynamics_joint} as follows:
\begin{equation}
 - {{\boldsymbol{n}}_j} + {{\boldsymbol{\tau}}_{j,{{min}}}} \leqslant \left[ {{{\boldsymbol{M}}_j}{{   }}-{\boldsymbol{J}}_j^T} \right]{\boldsymbol{\xi}} \leqslant  - {{\boldsymbol{n}}_j} + {{\boldsymbol{\tau}}_{j,{{max}}}},
\end{equation}
where $\boldsymbol{\tau}_{j,min}$ and $\boldsymbol{\tau}_{j,max} \in \mathbb{R}^{n_j}$ are the upper and lower bounds of the actuator torque, respectively.

\subsubsection{Friction constraints}
To avoid slipping, contact forces must be constrained within the friction cone, which is approximated by square pyramids:
$\left[ {{\boldsymbol{0}}\ {{\boldsymbol{C}}_{fr}}} \right]{\boldsymbol{\xi}} \leqslant {\boldsymbol{0}}$ with 
\begin{equation}
  {{\boldsymbol{C}}_{fr}} = \left[ {\begin{array}{*{20}{c}}
  {{{\boldsymbol{C}}_{lc}}}& \cdots &0 \\ 
   \vdots & \ddots & \vdots  \\ 
  0& \cdots &{{{\boldsymbol{C}}_{lc}}} 
\end{array}} \right],{{\boldsymbol{C}}_{lc}} = \left[ {\begin{array}{*{20}{l}}
  0&0&{ - 1} \\ 
  1&0&{ - \mu } \\ 
  { - 1}&0&{ - \mu } \\ 
  0&1&{ - \mu } \\ 
  0&{ - 1}&{ - \mu } 
\end{array}} \right].
\nonumber 
\end{equation}

\subsubsection{Torque computation}
With the optimal accelerations and contact forces solved from the hierarchical optimization, the torque commands for the actuators ${{\boldsymbol{\tau }}_j^d}$ can be computed using the actuated part of the dynamics \eqref{eq:dynamics_joint} as follows:
\begin{equation}
    {{\boldsymbol{\tau }}_j^d} = \left[ {{{\boldsymbol{M}}_j}{\text{  }}-{\boldsymbol{J}}_j^T} \right]{{\boldsymbol{\xi }}_d} + {{\boldsymbol{n}}_j}.
\end{equation}

\section{EXPERIMENTS}
Various validations have been conducted both in simulation and hardware. 
The humanoid robot used to validate the proposed approach is Orca \uppercase\expandafter{\romannumeral1} \cite{orca}, a human-sized robot, as shown in Fig.~\ref{fig:cover}. It features a total of 28 DoF, with 6 DoF for each manipulator, 6 DoF for each leg, 1 DoF for the waist, and 3 DoF for the head. The mass of the limbs accounts for approximately 64\% of the robot's total mass.

To simplify the control design, and without loss of generality, only the the manipulator and leg joints are controlled, while the head and waist joints are ignored and fixed. Therefore, the robot has $n_j = 24$ DoF in the experiment. The full control framework is illustrated in Fig.~\ref{fig:framework}, and all of the modules run on the user’s computer (Intel Core i7-13650HX@2.60GHz). The kino-dynamics planner (Sec.~\ref{Sec:MPC}) is solved by sequential quadratic programming (SQP) using the OCS2 toolbox\cite{OCS2} at a frequency of approximately 100Hz. The hierarchical optimization (Sec.~\ref{Sec:WBC}) is solved by QPOASES\cite{ferreau2014qpoases} at a frequency of approximately 800Hz. The state estimator fuses measurements from the motors’ encodes, IMU and Lidar based on the Kalman filter at a frequency of approximately 800Hz. The user input provides the desired pose and velocity of the robot's base at a frequency of approximately 30 Hz. The kinematics and dynamics of the robot is handled by the Pinocchio library\cite{carpentier2019pinocchio}. All the experiments discussed in this manuscript are supported by the video submission\footnote{Available at \url{https://youtu.be/--KyvVvE7XM}}.

\begin{figure*}[htp]
	\centering
	\subfloat[]{
        \label{fig:sub_sim_walk_vel}
		\includegraphics[width=0.32\linewidth]{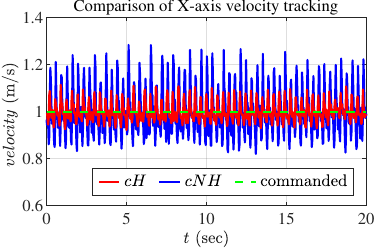}}
	\subfloat[]{
        \label{fig:sub_sim_walk_limbs}
		\includegraphics[width=0.32\linewidth]{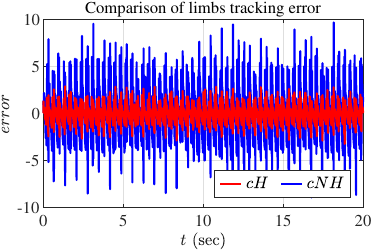}}
    \subfloat[]{
        \label{fig:sub_sim_walk_height}
		\includegraphics[width=0.32\linewidth]{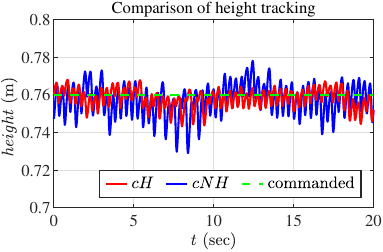}} \\
	\subfloat[]{
        \label{fig:sub_sim_walk_roll}
		\includegraphics[width=0.32\linewidth]{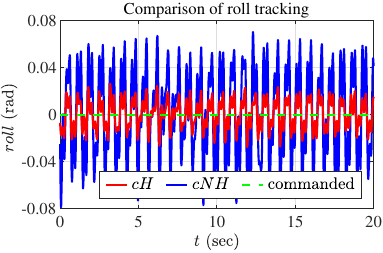}}  
    \subfloat[]{
        \label{fig:sub_sim_walk_pitch}
		\includegraphics[width=0.32\linewidth]{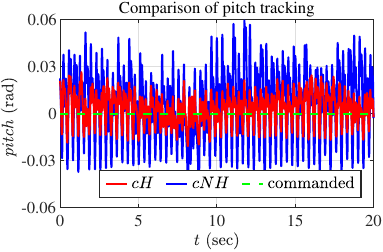}}
    \subfloat[]{
        \label{fig:sub_sim_walk_yaw}
		\includegraphics[width=0.32\linewidth]{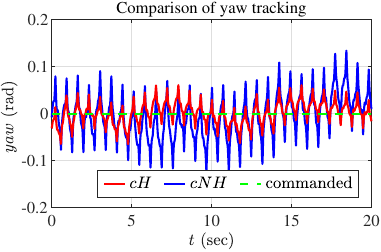}}
	\caption{Simulation results of dynamic walking with a desired velocity of 1.0~m/s. (a) is the velocity tracking response. The red solid line is the result caused by the {\textbf{cH}}, the blue solid line is the result caused by the {\textbf{cNH}}, and the green dotted line is the desired command. (b) is the limb tracking error. (c) is the height tracking response. (d), (e) and (f) are the orientation tracking response of roll, pitch, and yaw, respectively.}
\label{fig:sim_walk}
\end{figure*}

\subsection{Simulation Results for Comparison}
To validate the effectiveness of the proposed method, it was compared with the following two methods:

{\textbf{cI}}: This method, proposed in \cite{li2021force}, uses a MPC based on Single Rigid-Body Dynamics (SRBD) as the kino-dynamics planner, and employs inverse dynamics for tracking the reference motion and contact forces.

{\textbf{cNH}}: This method uses the kino-dynamics planner proposed in Sec.~\ref{Sec:MPC}, and employs quadratic programming (QP) from \cite{fahmi2019passive} to track the reference motion and contact forces. Unlike the approach proposed in the manuscript, it does not prioritize the tracking of limbs, base, and contact forces.

The method proposed in this manuscript is referred to as {\textbf{cH}}. All methods were compared and validated in a simulator developed in the Gazebo environment\cite{koenig2004design}. 

\subsubsection{Dynamic walking}
The first set of experiments focuses on the robot's balance in dynamic walking. The desire velocity was set to 0.5~m/s, 1.0~m/s and 1.2~m/s along the x-axis, respectively. From the video (from the 9th second to the 59th second), compared with \textbf{cI}, \textbf{cNH} and \textbf{cH} can generate coordinated upper limb motion, demonstrating the advantage of the kino-dynamics planner (Sec.~\ref{Sec:MPC}) proposed in the manuscript.

When the desire velocity was set to 0.5~m/s (from the 11th second to the 25th second), the robot controlled by \textbf{cI} fell. The reason for this is that during dynamic walking, the rapid motion of the limbs caused significant changes in the robot's mass and inertia distribution, increasing the model error in the SRBD of \textbf{cI}. This made it difficult to generate appropriate motion and contact forces to maintain the robot's balance.
When the desire velocity was set to 1.2~m/s (from the 42th second to the 59th second in the video), the robot controlled by \textbf{cNH} also fell. This occurred because, without hierarchical optimization, \textbf{cNH} exploited limb inertia to track the base references in underactuated directions, leading to unpredictable limb motions. This increased the limb tracking error, causing the robot to lose stability.

To quantify this issue, Fig.~\ref{fig:sim_walk} shows the results for \textbf{cH} and \textbf{cNH} when the desired velocity was set to 1.0~m/s. Due to the simplified system dynamics \eqref{eq:cd}, the roll angle of the humanoid robot's base is underactuated during dynamic walking. Compared to \textbf{cH},  \textbf{cNH} attempted to exploit the limbs' inertia to reduce the roll tracking error, which in turn increased the limb tracking error, as shown in Fig.\ref{fig:sub_sim_walk_roll} and Fig.\ref{fig:sub_sim_walk_limbs}. 
As the limb tracking error grew, \textbf{cNH} struggled to stabilize the robot, which is reflected in the increased velocity tracking error and base state tracking error, as shown in Fig.~\ref{fig:sub_sim_walk_vel}, Fig.~\ref{fig:sub_sim_walk_height}, Fig.~\ref{fig:sub_sim_walk_pitch} and Fig.~\ref{fig:sub_sim_walk_yaw}.

\begin{figure*}[tp]
	\centering
	\subfloat[]{
        \label{fig:sub_sim_disturb_x}
		\includegraphics[width=0.32\linewidth]{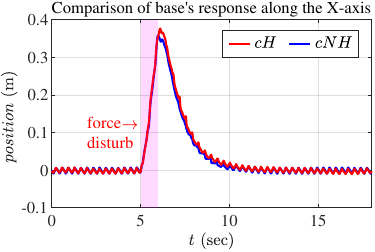}}
	\subfloat[]{
        \label{fig:sub_sim_disturb_y}
		\includegraphics[width=0.32\linewidth]{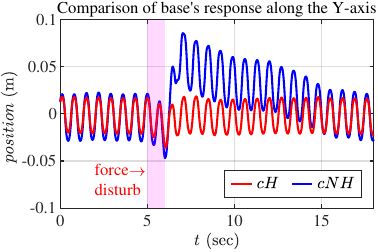}}
    \subfloat[]{
        \label{fig:sub_sim_disturb_height}
		\includegraphics[width=0.32\linewidth]{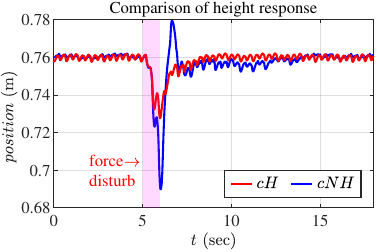}} \\
    \subfloat[]{
        \label{fig:sub_sim_disturb_roll}
		\includegraphics[width=0.32\linewidth]{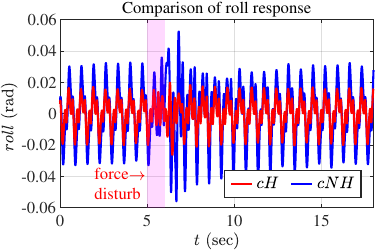}}
	\subfloat[]{
        \label{fig:sub_sim_disturb_pitch}
		\includegraphics[width=0.32\linewidth]{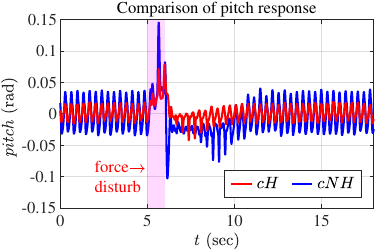}}  
    \subfloat[]{
        \label{fig:sub_sim_disturb_yaw}
		\includegraphics[width=0.32\linewidth]{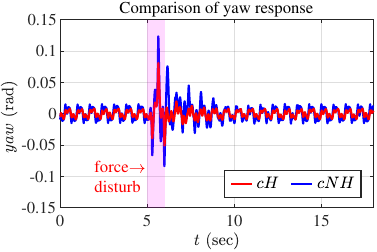}}
	\caption{Simulation results of disturbance rejection with an external force of 40~N applied to the base at the fifth second for 1 second.  (a) and (b) are the motion responses in the x-axis and y-axis directions, respectively. (c) is the motion responses in the height. (d), (e) and (f) are the orientation responses of roll, pitch, and yaw, respectively. 
The red solid line is the result caused by the {\textbf{cH}} and the blue solid line is the result caused by the {\textbf{cNH}}.
}
\label{fig:sim_disturbance}
\end{figure*}

\subsubsection{Disturbance rejection}
The second set of experiments focuses on the robot's balance with an external force disturbance. The external force was set to 20~N, 40~N and 60~N along the x-axis, respectively.

When the external force was set to 20~N (from the 60th second to the 80th second in the video), the robot controlled by \textbf{cI} fell. The reason is the SRBD struggles to generate the necessary motion and forces to maintain the balance of robots with heavy limbs.
When the external force was set to 60~N (from the 90th second to the 100th second in the video), the robot controlled by \textbf{cNH} also fell. The reason for this is that, although the kino-dynamics planner (Sec.~\ref{Sec:MPC}) leveraged the base-limbs coupling to maintain the robot's balance, the lack of hierarchical optimization in \textbf{cNH} led to increased tracking errors of the limbs. This resulted in unpredictable limb motions, ultimately disrupting the robot's balance.

Fig.~\ref{fig:sim_disturbance} presents the results with a 40~N external force applied to the base at the fifth second for 1 second. From Fig.~\ref{fig:sub_sim_disturb_y} and Fig.~\ref{fig:sub_sim_disturb_height}, the robot controlled by \textbf{cNH} exhibited deviation along the y-axis and oscillation in height after the external force was removed. This is due to the large control error in the limbs, which is amplified by inertial effects and transmitted to the base, causing it to deviate from the desired trajectory and making it difficult to maintain balance. In contrast, \textbf{cH} achieves better base-limb coupling, leading to more monotonic and smoother motion responses once the external forces are remove.

\begin{figure}[htbp]
	\centering
    \subfloat[]{
        \label{fig:sub_uneven_height}
		\includegraphics[width=0.47\linewidth]{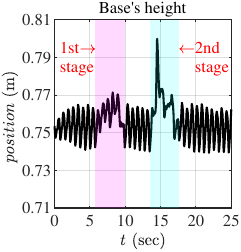}} 
    \subfloat[]{
        \label{fig:sub_uneven_yaw}
		\includegraphics[width=0.47\linewidth]{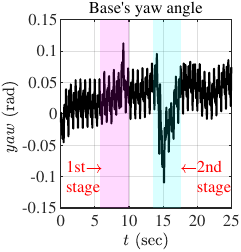}} \\
    \subfloat[]{
        \label{fig:sub_uneven_roll}
		\includegraphics[width=0.47\linewidth]{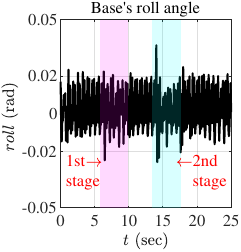}} 
	\subfloat[]{
        \label{fig:sub_uneven_pitch}
		\includegraphics[width=0.47\linewidth]{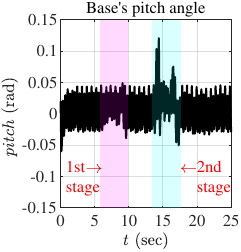}}  
	\caption{Hardware results of waking over uneven terrain. (a) is the motion response  in the height. (b), (c) and (d) are the orientation responses of yaw, pitch, and roll, respectively.
  }
\label{fig:sim_uneven}
\end{figure}

\subsection{Hardware Results}
The third set of experiments focuses on the performance of the hardware robot walking in real-world scenarios, including uneven terrain, outdoor environments, and resisting disturbances.
For the sake of space and without loss of generality, only the results from walking over uneven terrain are presented. Other experimental results can be found in the video (from the 120th second to the 180th second). The uneven terrain consists of one 5~cm and one 10~cm high stage. 
The motion response in the height and the orientation are shown in Fig.~\ref{fig:sim_uneven}.
As seen in Fig.~\ref{fig:sim_uneven}, between the 6th and 10th second, the robot successfully walked over the first stage, and between the 13th and 18th second, it walked the second stage. Using the approach proposed in this manuscript, the robot maintained stability while walking over stages of different heights. 
This demonstrates the robustness of the proposed approach in real-world scenarios.


\subsection{Real-time performance}
To demonstrate the effectiveness of the proposed approach in reducing computational complexity, the average computation time for the above experiments was recorded.

In the dynamic walking simulation experiment, the average computation times for the kino-dynamics planner and the hierarchical optimization were 6.85~ms and 0.58~ms, respectively. In the disturbance rejection simulation experiment, the average computation times for the kino-dynamics planner and the hierarchical optimization were 7.89~ms and 0.52~ms, respectively. In the hardware experiment, the average computation times for the kino-dynamics planner and the hierarchical optimization were 6.57~ms and 0.53~ms, respectively. These results demonstrate that by simplifying the approach, the proposed approach achieves a balance between performance and computational efficiency.

\section{CONCLUSIONS}
This paper presents a whole-body control framework for humanoid robots with heavy limbs, using a model-based approach. The proposed framework effectively mitigates the negative impact of heavy limbs on the robot's balance.
Through a well-designed model and constraints, the kino-dynamics planner in the framework generates reference motion and contact forces while accounting for base-limbs coupling in real-time.
Additionally, the hierarchical optimization effectively reduces limb control errors, ensuring compliance with the planner’s policy.

Experimental results demonstrate that the proposed framework achieves improved performance in various tasks. It enables dynamic walking at speeds of up to 1.2~m/s, responds to external disturbances of up to 60~N, and maintains balance on challenging terrains, such as uneven surfaces, and outdoor environments..

Future work will extend the proposed framework to enable human-like walking and other complex motions.

\addtolength{\textheight}{-12cm}   


\bibliographystyle{IEEEtran}
\bibliography{IEEEabrv,root}

\end{document}